\documentclass[11pt,a4paper]{article}
\usepackage[hyperref]{acl2020}
\usepackage{times}
\usepackage{latexsym}

\usepackage{microtype}
\usepackage{multirow}
\usepackage{hyperref}
\usepackage{url}
\usepackage{amsmath}
\usepackage{booktabs}
\usepackage{xcolor, colortbl}
\usepackage{graphicx}
\usepackage{longtable}

\definecolor{light-blue}{RGB}{138, 233, 255}

\aclfinalcopy 

\title{Towards a Deep Multi-layered Dialectal Language Analysis: A Case Study of African-American English}

\author{Jamell Dacon \\
  Michigan State University\\
  East Lansing, MI, USA \\
  \texttt{daconjam@msu.edu}}

\date{}

\begin{document}
\maketitle
\begin{abstract}
Currently, natural language processing (NLP) models proliferate language discrimination leading to potentially harmful societal impacts as a result of biased outcomes. For example, part-of-speech taggers trained on Mainstream American English (MAE) produce non-interpretable results when applied to African American English (AAE) as a result of language features not seen during training. In this work, we incorporate a human-in-the-loop paradigm to gain a better understanding of AAE speakers' behavior and their language use, and highlight the need for dialectal language inclusivity so that native AAE speakers can extensively interact with NLP systems while reducing feelings of disenfranchisement.

\end{abstract}

\section{Introduction}\label{sec:intro}

Over the years, social media users have leveraged online conversational platforms to perpetually express themselves online. 
{{For example}}, African American English (AAE)\footnote{A dialectal continuum previously known as Northern Negro English, Black English Vernacular (BEV), Black English, African American Vernacular English (AAVE), African American Language (AAL), Ebonics, and Non-standard English \citep{labov_1975, structure, green_2002, green_book, baugh2008linguistic, Bland-Stewart, linguistics-011619-030556}. It is often referred to as African American Language (AAL) and African American English (AAE). In this work, we use the denotation AAE.}, an English language variety is often heavily used on Twitter \cite{race_survey, language_survey}. This dialect continuum is neither spoken by \textit{all} African Americans or individuals who identify as BIPOC (Black, Indigenous, or People of Color), nor is it spoken \textit{only} by African Americans or BIPOC individuals \cite{race_survey, Bland-Stewart}. {{In some cases, AAE, a low-resource language (LRL) may be the first (or dominant) language, rather than the second (or non-dominant) language of an English speaker.}}

\begin{table*}[t]
\centering
\scalebox{0.8}{\begin{tabular}{c|c|l}
\hline
\multirow{2}{*}{\textbf{MAE}} & \textbf{Input} & I have never done this before \\ \cline{2-3} 
 & \textbf{Output} & (I, $<$PRP$>$), (have, $<$VBP$>$), (never, $<$RB$>$), (done, $<$VBN$>$), (that, $<$IN$>$), (before, $<$IN$>$)\\
\hline
\hline
\multirow{2}{*}{\textbf{AAE}} & \textbf{Input} & I \colorbox{light-blue}{aint} \colorbox{light-blue}{neva} did \colorbox{light-blue}{dat} \colorbox{light-blue}{befo}\\ \cline{2-3} 
 & \textbf{Output} & (I, $<${PRP}$>$), (aint, $<$\colorbox{pink}{VBP}$>$), (neva, $<$\colorbox{pink}{NN}$>$), (did, $<${VBD}$>$)(dat, $<$\colorbox{pink}{JJ}$>$), (befo, $<$\colorbox{pink}{NN}$>$)\\ \hline
\end{tabular}}
\caption{An illustrative example of POS tagging of semantically equivalent sentences written in MAE and AAE. Each \colorbox{light-blue}{blue} and \colorbox{pink}{red} highlight corresponds to linguistics features of AAE lexical items, and their misclassified NLTK (inferred) tags, respectively.}
\label{tab:pos_labels}
\end{table*}

Specifically, AAE is a regional dialect continuum that consists of a distinct set of lexical items, some of which have distinct semantic meanings, and may possess different syntactic structures/patterns than in Mainstream American English (MAE) (\textit{e.g.}, differentiating habitual \textit{be} and non-habitual \textit{be} usage) \citep{stewart-2014-now, dorn-2019-dialect, DescriptionAAE, race_survey, Bland-Stewart, baugh2008linguistic, language_survey, labov_1975}. In particular, \citet{green_2002} states that AAE possesses a morphologically invariant form of the verb that distinguishes between habitual action and currently occurring action, namely \textit{habitual \textbf{be}}. For example, ``the habitual be'' experiment\footnote{\url{https://www.umass.edu/synergy/fall98/ebonics3.html}} by University of Massachusetts Amherst’s Janice Jackson.

However, AAE is perceived to be ``bad english'' despite numerous studies by socio/raciolinguists and dialectologists in their attempts to quantify AAE as a legitimized language \cite{baugh2008linguistic, race_survey, Bland-Stewart, labov_1975}. 
\begin{quote}
    \small{``\textit{[T]he common misconception [is] that language use has primarily to do with words and what they mean. It doesn’t. It has primarily to do with people and what \textbf{they} mean.}'' \hfill -- \citet{Clark1992AskingQA}}
\end{quote}
\noindent Recently, online AAE has influenced the generation of resources for AAE-like text for natural language (NLP) and corpus linguistic tasks \textit{e.g.}, part-of-speech (POS) tagging \cite{jorgensen-etal-2016-learning, blodgett-etal-2018-twitter}, language generation \cite{groenwold-etal-2020-investigating} and automatic speech recognition \cite{dorn-2019-dialect, tatman17_interspeech}. POS tagging is a token-level text classification task where each token is assigned a corresponding word category label (see Table~\ref{tab:pos_labels}). It is an enabling tool for NLP applications such as a syntactic parsing, named entity recognition, corpus linguistics, etc. 
In this work, we incorporate a human-in-the-loop paradigm by directly involving affected (user) communities to understand context and word ambiguities in an attempt to study dialectal language inclusivity in NLP language technologies that are generally designed for dominant language varieties. \citet{Dacon_DGM} state that, 
\begin{quote}
    \small{``\textit{NLP systems aim to [learn] from natural language data, and mitigating social biases become a compelling matter not only for machine learning (ML) but for social justice as well.}'' }
\end{quote}

To address these issues, we aim to empirically study \textit{predictive bias} (see \citet{Swinton1981PREDICTIVEBI} for definition) \textit{i.e.}, if POS tagger models make predictions dependent on demographic language features, and attempt a dynamic approach in data-collection of non-standard spellings and lexical items. To examine the behaviors of AAE speakers and their language use, we first collect variable (morphological and phonological) rules of AAE language features from literature \cite{labov_1975, structure, green_2002, Bland-Stewart, stewart-2014-now, DBLP:journals/corr/BlodgettGO16, elazar-goldberg-2018-adversarial, baugh2008linguistic, green_book} (see Appendix~\ref{sp}). Then, we employ 5 trained sociolinguist Amazon Mechanical Turk (AMT) annotators\footnote{A HIT approval rate $\geq$ 95\% was used to select 5 bi-dialectal AMT annotators between the ages of 18 - 55, and completed $>$ 10,000 HITs and located within the United States.} who identify as bi-dialectal dominant AAE speakers to address the issue of lexical, semantic and syntactic ambiguity of tweets (see Appendix~\ref{sec:human} for annotation guidelines). Next, we incorporate a human-in-the-loop paradigm by recruiting 20 crowd-sourced diglossic annotators to evaluate AAE language variety (see Table \ref{tab:category}). Finally, we conclude by expanding on the need for dialectal language inclusivity.

\section{Related Work }\label{sec:related}

Previous works regarding AAE linguistic features have analyzed tasks such as unsupervised domain adaptation for AAE-like language \citep{jorgensen-etal-2016-learning}, detecting AAE syntax\citep{stewart-2014-now}, language identification \cite{DBLP:journals/corr/BlodgettO17}, voice recognition and transcription \cite{dorn-2019-dialect}, dependency parsing \citep{blodgett-etal-2018-twitter}, dialogue systems \citep{liu-etal-2020-gender}, hate speech/toxic language detection and examining racial bias \citep{sap-etal-2019-risk, 10.1145/3465416.3483299, xia-etal-2020-demoting, DBLP:journals/corr/abs-2005-13041, zhou-etal-2021-challenges, moza, xu-etal-2021-detoxifying, doi:10.1073/pnas.1915768117}, and language generation \citep{groenwold-etal-2020-investigating}. These central works are conclusive for highlighting systematic biases of natural language processing (NLP) systems when employing AAE in common downstream tasks.  

Although we mention popular works incorporating AAE, this dialectal continuum has been largely ignored and underrepresented by the NLP community in comparison to MAE. Such lack of language diversity cases constitutes technological inequality to minority groups, for example, by African Americans or BIPOC individuals, and may intensify feelings of disenfranchisement due to monolingualism. We refer to this pitfall as the \textit{inconvenient truth} \textit{i.e.},
\begin{quote}
    \small{``\textit{[I]f the systems show discriminatory behaviors in the interactions, the user experience will be adversely affected.}''} \hfill--- \citet{liu-etal-2020-gender}
 \end{quote} 
\noindent Therefore, we define fairness as the model's ability to correctly predict each tag while performing zero-shot transfer via dialectal language inclusivity. 

Moreover, these aforementioned works do not discuss nor reflect on the ``\textit{role of the speech and language technologies in sustaining language use}'' \citep{labov_1975, bird-2020-decolonising, language_survey} as,
\begin{quote}
    \small{``\textit{... models are expected to make predictions with the semantic information rather than with the demographic group identity information}''} \hfill--- \citet{zhang-etal-2020-demographics}.
\end{quote}
Interactions with everyday items is increasingly mediated through language, yet systems have limited ability to process less-represented dialects such as AAE. For example, a common AAE phrase, ``\textit{I had a long \textbf{ass} day}'' would receive a lower sentiment polarity score because of the word ``\textit{\textbf{ass}}'', a (noun) term typically classified as offensive; however, in AAE, this term is often used as an emphatic, cumulative adjective and perceived as non-offensive. \\
\begin{figure*}[t]
    \centering
    \includegraphics[scale=0.34]{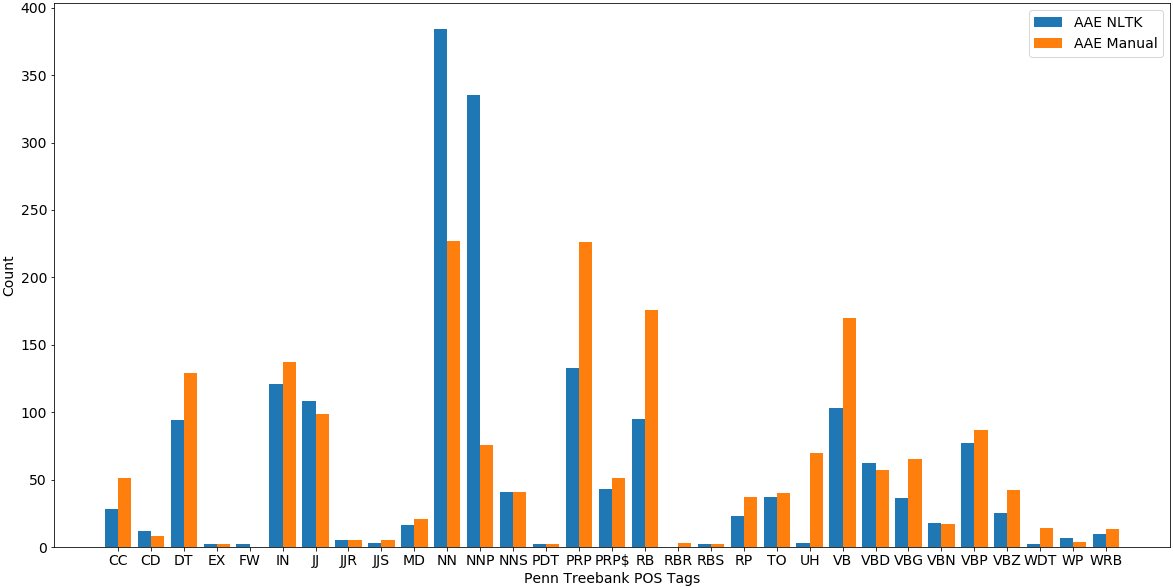}
    \caption{An illustration of inferred and manually-annotated AAE tag counts from $k$ randomly sampled tweets.}
    \label{fig:AAE_POS_Tags}
\end{figure*}

\noindent \textbf{Motivation:} We want to test our hypothesis that training each model on correctly tagged AAE language features will improve the model’s performance, interpretability, explainability, and usability to reduce predictive bias.

\section{Dataset and Annotation}\label{sec:data}

\begin{table*}[ht]
\centering
\scalebox{0.85}{\begin{tabular}{l|l|l|l}
\hline
\textbf{Tags} & \textbf{Category} &  \textbf{AAE Example(s)} & \textbf{MAE Equivalent(s)} \\ \hline
CC & Coordinating Conjunction &   \textit{doe/tho, n, bt} & \textit{though, and, but} \\
DT & Determiner &   \textit{da, dis, dat} & \textit{the, this, that} \\ 
EX & Existential There &   \textit{dea} & \textit{there} \\ 
IN & Preposition/ Conjunction &   \textit{fa, cuz/cause, den} & \textit{for, because, than} \\ 
JJ & Adjective &   \textit{foine, hawt} & \textit{fine, hot} \\ 
PRP & Pronoun &   \textit{u, dey, dem} & \textit{you, they, them} \\ 
PRP\$ & Personal Pronoun &   \textit{ha} & \textit{her} \\ 
RB & {Adverb} &   \textit{tryna, finna, jus} & \textit{trying to, fixing to, just} \\ 
RBR & {Adverb, comparative} &   \textit{mo, betta, hotta} & \textit{more, better, hotter} \\ 
RP & {Particle} &   \textit{bout, thru} & \textit{about, through} \\ 
TO & {Infinite marker} &   \textit{ta} & \textit{to} \\ 
UH & {Interjection} &   \textit{wassup, ion, ian} & \textit{what's up, I don't} \\ 
VBG & Verb, gerund~ &   \textit{sleepin, gettin} & \textit{sleeping, getting} \\ 
VBZ & Verb, 3rd-person present tense &   \textit{iz} & \textit{is} \\ 
WDT & {Wh-determiner} &   \textit{dat, wat, wus, wen} & \textit{that, what, what's, when} \\ 
WRB & {Wh-adverb} &   \textit{hw} & \textit{how} \\ \hline
\end{tabular}}
\caption{Accurately tagged (observed) AAE and English phonological and morphological \textbf{linguistic} feature(s) accompanied by their respective MAE equivalent(s).} \label{tab:category}
\end{table*}

\subsection{Dataset}
We collect 3000 demographically-aligned African American (AA) tweets possessing an average of 7 words per tweet from the publicly available TwitterAAE corpus by \citet{DBLP:journals/corr/BlodgettGO16}. Each tweet is accompanied by inferred geolocation topic model probabilities from Twitter + Census demographics and word likelihoods to calculate demographic dialect proportions. We aim to minimize (linguistic) discrimination by sampling tweets that possess over 99\% confidence to develop ``fair'' NLP tools that are originally designed for dominant language varieties by integrating non-standardized varieties. More information about the TwitterAAE dataset, including its statistical information, annotation process, and the link(s) to downloadable versions can be found in Appendix~\ref{dataset}.

\subsection{Preprocessing}\label{sec:preprocessing}

As it is common for most words on social media to be plausibly semantically equivalent, we denoise each tweet as tweets typically possess unusual spelling patterns, repeated letter, emoticons and emojis\footnote{Emoticons are particular textual features made of punctuation such as exclamation marks, letters, and/or numbers to create pictorial icons to display an emotion or sentiment (\textit{e.g.}, ``;)'' $\Rightarrow$ \textit{winking smile}), while emojis are small text-like pictographs of faces, objects, symbols, etc.}. We replace sequences of multiple repeated letters with three repeated letters (\textit{e.g.}, \textit{Hmmmmmmmm} $\rightarrow$ \textit{Hmmm}), and remove all punctuation, ``\texttt{@}'' handles of users and emojis. Essentially, we aim to denoise each tweet only to capture non-standard spellings and lexical items more efficiently.

\subsection{Annotation}\label{sec:annotation}

First, we employ off-the-shelf taggers such as {spacy}\footnote{\url{https://spacy.io}} and {TwitterNLP}\footnote{\url{https://github.com/ianozsvald/ark-tweet-nlp-python}}; however, the Natural Language Toolkit (NLTK) \citep{loper-bird-2002-nltk} provides a more fine-grained Penn Treebank Tagset (PTB)\footnote{\url{https://www.guru99.com/pos-tagging-chunking-nltk.html}} along with evaluation metrics per tag such as F1 score. Next, we focus on aggregating the appropriate tags by collecting and manually-annotating tags from AAE/slang-specific dictionaries to assist the AMT annotators, and later we contrast these aggregated tags with inferred NLTK PTB inferred tags. In  Figure~\ref{fig:AAE_POS_Tags}, we display NLTK inferred and manually-annotated AAE tags from $k = 300$ randomly sampled tweets.
\begin{itemize}
    \item {\textbf{The Online Slang Dictionary}} (American, English, and Urban slang)\footnote{\url{http://onlineslangdictionary.com}} -  created in 1996, this is the oldest web dictionary of slang words, neologisms, idioms, aphorisms, jargon, informal speech, and figurative usages. This dictionary possesses more than 24,000 real definitions and tags for over 17,000 slang words and phrases, 600 categories of meaning, word use mapping and aids in addressing lexical ambiguity.
    
    \item \textbf{Word Type}\footnote{\url{https://wordtype.org/}} - an open source POS focused dictionary of words based on the Wiktionary\footnote{\url{https://www.wiktionary.org}} project by Wikimedia\footnote{\url{https://www.wikimedia.org}}. Researchers have parsed Wiktionary and other sources, including real definitions and categorical POS word use cases necessary to address the issue of lexical, semantic and syntactic ambiguity.
\end{itemize}

\subsection{Human Evaluation} After an initial training of the AMT annotators, we task each annotator to annotate each tweet with the appropriate POS tags. Then, as a calibration study we attempt to measure the inter-annotator agreement (IAA) using Krippendorff's $\alpha$. By using NLTK’s \cite{loper-bird-2002-nltk} $\mathrm{nltk.metrics.agreement}$, we calculate a Krippendorf's $\alpha$ of 0.88. We did not observe notable distinctions in annotator agreement across the individual tweets. We later randomly sampled 300 annotated tweets and recruit 20 crowd-sourced annotators to evaluate AAE language variety.
To recruit 20 diglossic annotators\footnote{Note that we did not collect certain demographic information such as gender or race, only basic demographics such as age (18-55 years), state and country of residence.}, we created a volunteer questionnaire with annotation guildlines, and released it on LinkedIn. The full annotation guildlines can be found in Appendix~\ref{sec:human}. 
Each recruited annotator is tasked to judge sampled tweets and list their MAE equivalents to examine contextual differences of simple, deterministic morphosyntactic substitutions of dialect-specific vocabulary in standard English or MAE texts---a \textit{reverse} study to highlight several varieties of AAE (see Table \ref{tab:category}).
\section{Methodology}\label{sec:methods}

In this section, we describe our approach to perform a preliminary study to validate the existence of predictive bias \cite{elazar-goldberg-2018-adversarial, shah-etal-2020-predictive} in POS models. We first introduce the POS tagging, and then propose two ML sequence models.

\subsection{Part-of-Speech (POS) Tagging}\label{sec:POS_Tagging}

We consider POS tagging as it represents word syntactic categories and serves as a pre-annotation tool for numerous downstream tasks, especially for non-standardized English language varieties such as AAE \citep{zampieri_nakov_scherrer_2020}. 
Common tags include prepositions, adjective, pronoun, noun, adverb, verb, interjection, etc., where multiple POS tags can be assigned to particular words due to syntactic structural patterns. This can also lead to misclassification of non-standardized words that do not exist in popular pre-trained NLP models.

\subsection{Models}\label{sec:models}

We propose to implement two well known sequence modeling algorithms, namely a Bidirectional Long Short Term Memory (Bi-LSTM) network, a deep neutral network (DNN) \citep{10.1162/neco.1997.9.8.1735, GRAVES2005602} that has been used for POS tagging \citep{ling-etal-2015-finding, plank-etal-2016-multilingual}, and a Conditional Random Field (CRF) \cite{laffertyconditional} typically used to identify entities or patterns in texts by exploiting previously learned word data. \\

\noindent \textbf{Taggers:} First, we use NLTK \citep{loper-bird-2002-nltk} for automatic tagging; then, we pre-define a feature function for our CRF model where we optimized its L1 and L2 regularization parameters to 0.25 and 0.3, respectively. Later, we train our Bi-LSTM network for 40 epochs with an Adam optimizer, and a learning rate of 0.001. Note that each model would be accompanied by error analysis for a 70-30 split of the data with 5-fold cross-validation to obtain model classification reports, for metrics such as precision, recall and F1-score.

\section{Operationalization of AAE as an English Language Variety}\label{sec:discussion}

As (online) AAE can incorporate non-standardized spellings and lexical items, there is an active need for a human-in-the-loop paradigm as humans provide various forms of feedback in different stages of workflow. This can significantly improve the model’s performance, interpretability, explainability, and usability.
Therefore, crowd-sourcing to develop language technologies that consider who created the data will lead to the inclusion of diverse training data, and thus, decrease feelings of marginalization. For example, {CORAAL}\footnote{\url{https://oraal.uoregon.edu/coraal}}, is an online resource that features AAL text data, recorded speech data, etc., into new and existing NLP technologies, AAE speakers can extensively interact with current NLP language technologies.

Consequently, to quantitatively and qualitatively ensure fairness in NLP tools, artificial intelligence (AI) and NLP researchers need to go beyond evaluation measures, word definitions and word order to assess AAE on a token-level to better understand context, culture and word ambiguities. We encourage both AI and NLP practitioners to prioritize collecting a set of relevant labeled training data with several examples of informal phrases, expressions, idioms, and regional-specific varieties. Specifically, in models intended for broad use such as sentiment analysis by partnering with low-resource and dialectal communities to develop impactful speech and language technologies for dialect continua such as AAE to minimize further stigmatization of an already stigmatized minority group. 

\section{Conclusion}\label{sec:conclusion}

Throughout this work, we highlight the need to develop language technologies for such varieties, pushing back against potentially discriminatory practices (in many cases, discriminatory through oversight more than malice). Our work calls for NLP researchers to consider both social and racial hierarchies sustained or intensified by current computational linguistic research. By shifting towards a human-in-the-loop paradigm to conduct deep multi-layered dialectal language analysis of AAE to counter-attack erasure and several forms of biases such as \textit{selection bias}, \textit{label bias}, \textit{model overamplification}, and \textit{semantic bias} (see \citet{shah-etal-2020-predictive} for definitions) in NLP.

We hope our dynamic approach can encourage practitioners, researchers and developers for AAE inclusive work, and that our contributions can pave the way for normalizing the use of a human-in-the-loop paradigm both to obtain new data and create NLP tools to better comprehend  underrepresented dialect continua and English language varieties. In this way, NLP community can revolutionize the ways in which humans and technology cooperate by considering certain demographic attributes such as culture, background, race and gender when developing and deploying NLP models.

\section{Limitations And Ethical Considerations}\label{sec:ethics}

All authors must warrant that increased model performance for non-standard varieties such as underrepresented dialects, non-standard spellings or lexical items in NLP systems can potentially enable automated discrimination. In this work, we \textit{solely} attempt to highlight the need for dialectal inclusivity for the development of impactful speech and language technologies in the future, and do not intend for increased feelings of marginalization of an already stigmatized community.

\section{Acknowledgements}\label{sec:ack}

The authors would like to thank Shaylnn L.A. Crum-Dacon, Serena Lotreck, Brianna Brown and Kenia Segura Abá, Jyothi Kumar and Shin-Han Shiu for their support and the anonymous reviewers for their constructive comments. 

\bibliography{main}

\begin{thebibliography}{41}
\expandafter\ifx\csname natexlab\endcsname\relax\def\natexlab#1{#1}\fi

\bibitem[{Bailey et~al.(1998)Bailey, Baugh, Mufwene, and Rickford}]{structure}
Guy Bailey, John Baugh, Salikoko~S. Mufwene, and John~R. Rickford. 1998.
\newblock \href {https://doi.org/10.4324/9780203355596} {\emph{African-American
  English: Structure, History and Use (1st ed.)}}.
\newblock Routledge.

\bibitem[{Baugh(2008)}]{baugh2008linguistic}
John Baugh. 2008.
\newblock Linguistic discrimination.
\newblock In \emph{1. Halbband}, pages 709--714. De Gruyter Mouton.

\bibitem[{Bird(2020)}]{bird-2020-decolonising}
Steven Bird. 2020.
\newblock \href {https://doi.org/10.18653/v1/2020.coling-main.313}
  {Decolonising speech and language technology}.
\newblock In \emph{Proceedings of the 28th International Conference on
  Computational Linguistics}, pages 3504--3519, Barcelona, Spain (Online).
  International Committee on Computational Linguistics.

\bibitem[{Bland-Stewart(2005)}]{Bland-Stewart}
Linda~M. Bland-Stewart. 2005.
\newblock Difference or deficit in speakers of african american english?
\newblock
  \url{https://leader.pubs.asha.org/doi/10.1044/leader.FTR1.10062005.6}.

\bibitem[{Blodgett et~al.(2020)Blodgett, Barocas, Daum{\'e}~III, and
  Wallach}]{language_survey}
Su~Lin Blodgett, Solon Barocas, Hal Daum{\'e}~III, and Hanna Wallach. 2020.
\newblock \href {https://doi.org/10.18653/v1/2020.acl-main.485} {Language
  (technology) is power: A critical survey of {``}bias{''} in {NLP}}.
\newblock In \emph{Proceedings of the 58th Annual Meeting of the Association
  for Computational Linguistics}, pages 5454--5476, Online. Association for
  Computational Linguistics.

\bibitem[{Blodgett et~al.(2016)Blodgett, Green, and
  O{'}Connor}]{DBLP:journals/corr/BlodgettGO16}
Su~Lin Blodgett, Lisa Green, and Brendan O{'}Connor. 2016.
\newblock \href {https://doi.org/10.18653/v1/D16-1120} {Demographic dialectal
  variation in social media: A case study of {A}frican-{A}merican {E}nglish}.
\newblock In \emph{Proceedings of the 2016 Conference on Empirical Methods in
  Natural Language Processing}, pages 1119--1130, Austin, Texas. Association
  for Computational Linguistics.

\bibitem[{Blodgett and O'Connor(2017)}]{DBLP:journals/corr/BlodgettO17}
Su~Lin Blodgett and Brendan O'Connor. 2017.
\newblock \href {http://arxiv.org/abs/1707.00061} {Racial disparity in natural
  language processing: {A} case study of social media african-american
  english}.
\newblock \emph{CoRR}, abs/1707.00061.

\bibitem[{Blodgett et~al.(2018)Blodgett, Wei, and
  O{'}Connor}]{blodgett-etal-2018-twitter}
Su~Lin Blodgett, Johnny Wei, and Brendan O{'}Connor. 2018.
\newblock \href {https://doi.org/10.18653/v1/P18-1131} {{T}witter {U}niversal
  {D}ependency parsing for {A}frican-{A}merican and mainstream {A}merican
  {E}nglish}.
\newblock In \emph{Proceedings of the 56th Annual Meeting of the Association
  for Computational Linguistics (Volume 1: Long Papers)}, pages 1415--1425,
  Melbourne, Australia. Association for Computational Linguistics.

\bibitem[{Clark and Schober(1992)}]{Clark1992AskingQA}
Herbert~H. Clark and Michael~F. Schober. 1992.
\newblock Asking questions and influencing answers.
\newblock In \emph{Russell Sage Foundation}.

\bibitem[{Dacon and Liu(2021)}]{Dacon_DGM}
Jamell Dacon and Haochen Liu. 2021.
\newblock \href {https://doi.org/10.1145/3442442.3452325} {Does gender matter
  in the news? detecting and examining gender bias in news articles}.
\newblock In \emph{Companion Proceedings of the Web Conference 2021}, WWW '21,
  page 385–392, New York, NY, USA. Association for Computing Machinery.

\bibitem[{Davidson and Bhattacharya(2020)}]{DBLP:journals/corr/abs-2005-13041}
Thomas Davidson and Debasmita Bhattacharya. 2020.
\newblock \href {https://arxiv.org/abs/2005.13041} {Examining racial bias in an
  online abuse corpus with structural topic modeling}.
\newblock \emph{CoRR}, abs/2005.13041.

\bibitem[{Dorn(2019)}]{dorn-2019-dialect}
Rachel Dorn. 2019.
\newblock \href {https://doi.org/10.26615/issn.2603-2821.2019_003}
  {Dialect-specific models for automatic speech recognition of {A}frican
  {A}merican {V}ernacular {E}nglish}.
\newblock In \emph{Proceedings of the Student Research Workshop Associated with
  RANLP 2019}, pages 16--20, Varna, Bulgaria. INCOMA Ltd.

\bibitem[{Elazar and Goldberg(2018)}]{elazar-goldberg-2018-adversarial}
Yanai Elazar and Yoav Goldberg. 2018.
\newblock \href {https://doi.org/10.18653/v1/D18-1002} {Adversarial removal of
  demographic attributes from text data}.
\newblock In \emph{Proceedings of the 2018 Conference on Empirical Methods in
  Natural Language Processing}, pages 11--21, Brussels, Belgium. Association
  for Computational Linguistics.

\bibitem[{Field et~al.(2021)Field, Blodgett, Waseem, and
  Tsvetkov}]{race_survey}
Anjalie Field, Su~Lin Blodgett, Zeerak Waseem, and Yulia Tsvetkov. 2021.
\newblock \href {http://arxiv.org/abs/2106.11410} {A survey of race, racism,
  and anti-racism in {NLP}}.
\newblock \emph{CoRR}, abs/2106.11410.

\bibitem[{Graves and Schmidhuber(2005)}]{GRAVES2005602}
Alex Graves and Jürgen Schmidhuber. 2005.
\newblock \href {https://doi.org/https://doi.org/10.1016/j.neunet.2005.06.042}
  {Framewise phoneme classification with bidirectional lstm and other neural
  network architectures}.
\newblock \emph{Neural Networks}, 18(5):602--610.
\newblock IJCNN 2005.

\bibitem[{Green(2014)}]{green_book}
Jonathon Green. 2014.
\newblock \emph{The vulgar tongue: Green’s history of slang.}
\newblock Oxford University Press, New York, USA.

\bibitem[{Green(2002)}]{green_2002}
Lisa~J. Green. 2002.
\newblock \href {https://doi.org/10.1017/CBO9780511800306} {\emph{African
  American English: A Linguistic Introduction}}.
\newblock Cambridge University Press.

\bibitem[{Groenwold et~al.(2020)Groenwold, Ou, Parekh, Honnavalli, Levy, Mirza,
  and Wang}]{groenwold-etal-2020-investigating}
Sophie Groenwold, Lily Ou, Aesha Parekh, Samhita Honnavalli, Sharon Levy, Diba
  Mirza, and William~Yang Wang. 2020.
\newblock \href {https://doi.org/10.18653/v1/2020.emnlp-main.473}
  {Investigating {A}frican-{A}merican {V}ernacular {E}nglish in
  transformer-based text generation}.
\newblock In \emph{Proceedings of the 2020 Conference on Empirical Methods in
  Natural Language Processing (EMNLP)}, pages 5877--5883, Online. Association
  for Computational Linguistics.

\bibitem[{Halevy et~al.(2021)Halevy, Harris, Bruckman, Yang, and
  Howard}]{10.1145/3465416.3483299}
Matan Halevy, Camille Harris, Amy Bruckman, Diyi Yang, and Ayanna Howard. 2021.
\newblock \href {https://doi.org/10.1145/3465416.3483299} {Mitigating racial
  biases in toxic language detection with an equity-based ensemble framework}.
\newblock New York, NY, USA. Association for Computing Machinery.

\bibitem[{Hochreiter and Schmidhuber(1997)}]{10.1162/neco.1997.9.8.1735}
Sepp Hochreiter and J\"{u}rgen Schmidhuber. 1997.
\newblock \href {https://doi.org/10.1162/neco.1997.9.8.1735} {Long short-term
  memory}.
\newblock \emph{Neural Comput.}, 9(8):1735–1780.

\bibitem[{Jones(2015)}]{DescriptionAAE}
Taylor Jones. 2015.
\newblock \href {https://doi.org/10.1215/00031283-3442117} {Toward a
  description of african american vernacular english dialect regions using
  “black twitter”}.
\newblock \emph{American Speech}, 90:403--440.

\bibitem[{J{\o}rgensen et~al.(2016)J{\o}rgensen, Hovy, and
  S{\o}gaard}]{jorgensen-etal-2016-learning}
Anna J{\o}rgensen, Dirk Hovy, and Anders S{\o}gaard. 2016.
\newblock \href {https://doi.org/10.18653/v1/N16-1130} {Learning a {POS} tagger
  for {AAVE}-like language}.
\newblock In \emph{Proceedings of the 2016 Conference of the North {A}merican
  Chapter of the Association for Computational Linguistics: Human Language
  Technologies}, pages 1115--1120, San Diego, California. Association for
  Computational Linguistics.

\bibitem[{King(2020)}]{linguistics-011619-030556}
Sharese King. 2020.
\newblock \href {https://doi.org/10.1146/annurev-linguistics-011619-030556}
  {From african american vernacular english to african american language:
  Rethinking the study of race and language in african americans’ speech}.
\newblock \emph{Annual Review of Linguistics}, 6(1):285--300.

\bibitem[{Koenecke et~al.(2020)Koenecke, Nam, Lake, Nudell, Quartey, Mengesha,
  Toups, Rickford, Jurafsky, and Goel}]{doi:10.1073/pnas.1915768117}
Allison Koenecke, Andrew Nam, Emily Lake, Joe Nudell, Minnie Quartey, Zion
  Mengesha, Connor Toups, John~R. Rickford, Dan Jurafsky, and Sharad Goel.
  2020.
\newblock \href {https://doi.org/10.1073/pnas.1915768117} {Racial disparities
  in automated speech recognition}.
\newblock \emph{Proceedings of the National Academy of Sciences},
  117(14):7684--7689.

\bibitem[{Labov(1975)}]{labov_1975}
William Labov. 1975.
\newblock \href {https://doi.org/10.1017/S0047404500004668} {Ralph fasold,
  tense marking in black english: a linguistic and social analysis. washington,
  d.c.: Center for applied linguistics, 1972. pp. 254.}
\newblock \emph{Language in Society}, 4(2):222–227.

\bibitem[{Lafferty et~al.()Lafferty, McCallum, and
  Pereira}]{laffertyconditional}
John Lafferty, Andrew McCallum, and Fernando~CN Pereira.
\newblock Conditional random fields: Probabilistic models for segmenting and
  labeling sequence data.

\bibitem[{Ling et~al.(2015)Ling, Dyer, Black, Trancoso, Fermandez, Amir,
  Marujo, and Lu{\'\i}s}]{ling-etal-2015-finding}
Wang Ling, Chris Dyer, Alan~W Black, Isabel Trancoso, Ram{\'o}n Fermandez,
  Silvio Amir, Lu{\'\i}s Marujo, and Tiago Lu{\'\i}s. 2015.
\newblock \href {https://doi.org/10.18653/v1/D15-1176} {Finding function in
  form: Compositional character models for open vocabulary word
  representation}.
\newblock In \emph{Proceedings of the 2015 Conference on Empirical Methods in
  Natural Language Processing}, pages 1520--1530, Lisbon, Portugal. Association
  for Computational Linguistics.

\bibitem[{Liu et~al.(2020)Liu, Dacon, Fan, Liu, Liu, and
  Tang}]{liu-etal-2020-gender}
Haochen Liu, Jamell Dacon, Wenqi Fan, Hui Liu, Zitao Liu, and Jiliang Tang.
  2020.
\newblock \href {https://doi.org/10.18653/v1/2020.coling-main.390} {Does gender
  matter? towards fairness in dialogue systems}.
\newblock In \emph{Proceedings of the 28th International Conference on
  Computational Linguistics}, pages 4403--4416, Barcelona, Spain (Online).
  International Committee on Computational Linguistics.

\bibitem[{Loper and Bird(2002)}]{loper-bird-2002-nltk}
Edward Loper and Steven Bird. 2002.
\newblock \href {https://doi.org/10.3115/1118108.1118117} {{NLTK}: The natural
  language toolkit}.
\newblock In \emph{Proceedings of the {ACL}-02 Workshop on Effective Tools and
  Methodologies for Teaching Natural Language Processing and Computational
  Linguistics}, pages 63--70, Philadelphia, Pennsylvania, USA. Association for
  Computational Linguistics.

\bibitem[{Mozafari et~al.(2020)Mozafari, Farahbakhsh, and Crespi}]{moza}
Marzieh Mozafari, Reza Farahbakhsh, and Noël Crespi. 2020.
\newblock \href {https://doi.org/10.1371/journal.pone.0237861} {Hate speech
  detection and racial bias mitigation in social media based on bert model}.
\newblock \emph{PLOS ONE}, 15:1--26.

\bibitem[{Plank et~al.(2016)Plank, S{\o}gaard, and
  Goldberg}]{plank-etal-2016-multilingual}
Barbara Plank, Anders S{\o}gaard, and Yoav Goldberg. 2016.
\newblock \href {https://doi.org/10.18653/v1/P16-2067} {Multilingual
  part-of-speech tagging with bidirectional long short-term memory models and
  auxiliary loss}.
\newblock In \emph{Proceedings of the 54th Annual Meeting of the Association
  for Computational Linguistics (Volume 2: Short Papers)}, pages 412--418,
  Berlin, Germany. Association for Computational Linguistics.

\bibitem[{Sap et~al.(2019)Sap, Card, Gabriel, Choi, and
  Smith}]{sap-etal-2019-risk}
Maarten Sap, Dallas Card, Saadia Gabriel, Yejin Choi, and Noah~A. Smith. 2019.
\newblock \href {https://doi.org/10.18653/v1/P19-1163} {The risk of racial bias
  in hate speech detection}.
\newblock In \emph{Proceedings of the 57th Annual Meeting of the Association
  for Computational Linguistics}, pages 1668--1678, Florence, Italy.
  Association for Computational Linguistics.

\bibitem[{Shah et~al.(2020)Shah, Schwartz, and
  Hovy}]{shah-etal-2020-predictive}
Deven~Santosh Shah, H.~Andrew Schwartz, and Dirk Hovy. 2020.
\newblock \href {https://doi.org/10.18653/v1/2020.acl-main.468} {Predictive
  biases in natural language processing models: A conceptual framework and
  overview}.
\newblock In \emph{Proceedings of the 58th Annual Meeting of the Association
  for Computational Linguistics}, pages 5248--5264, Online. Association for
  Computational Linguistics.

\bibitem[{Stewart(2014)}]{stewart-2014-now}
Ian Stewart. 2014.
\newblock \href {https://doi.org/10.3115/v1/E14-3004} {Now we stronger than
  ever: {A}frican-{A}merican {E}nglish syntax in {T}witter}.
\newblock In \emph{Proceedings of the Student Research Workshop at the 14th
  Conference of the {E}uropean Chapter of the Association for Computational
  Linguistics}, pages 31--37, Gothenburg, Sweden. Association for Computational
  Linguistics.

\bibitem[{Swinton(1981)}]{Swinton1981PREDICTIVEBI}
Spencer~S. Swinton. 1981.
\newblock Predictive bias in graduate admissions tests.
\newblock \emph{ETS Research Report Series}, 1981.

\bibitem[{Tatman and Kasten(2017)}]{tatman17_interspeech}
Rachael Tatman and Conner Kasten. 2017.
\newblock \href {https://doi.org/10.21437/Interspeech.2017-1746} {{Effects of
  Talker Dialect, Gender \& Race on Accuracy of Bing Speech and YouTube
  Automatic Captions}}.
\newblock In \emph{Proc. Interspeech 2017}, pages 934--938.

\bibitem[{Xia et~al.(2020)Xia, Field, and Tsvetkov}]{xia-etal-2020-demoting}
Mengzhou Xia, Anjalie Field, and Yulia Tsvetkov. 2020.
\newblock \href {https://doi.org/10.18653/v1/2020.socialnlp-1.2} {Demoting
  racial bias in hate speech detection}.
\newblock In \emph{Proceedings of the Eighth International Workshop on Natural
  Language Processing for Social Media}, pages 7--14, Online. Association for
  Computational Linguistics.

\bibitem[{Xu et~al.(2021)Xu, Pathak, Wallace, Gururangan, Sap, and
  Klein}]{xu-etal-2021-detoxifying}
Albert Xu, Eshaan Pathak, Eric Wallace, Suchin Gururangan, Maarten Sap, and Dan
  Klein. 2021.
\newblock \href {https://doi.org/10.18653/v1/2021.naacl-main.190} {Detoxifying
  language models risks marginalizing minority voices}.
\newblock In \emph{Proceedings of the 2021 Conference of the North American
  Chapter of the Association for Computational Linguistics: Human Language
  Technologies}, pages 2390--2397, Online. Association for Computational
  Linguistics.

\bibitem[{Zampieri et~al.(2020)Zampieri, Nakov, and
  Scherrer}]{zampieri_nakov_scherrer_2020}
Marcos Zampieri, Preslav Nakov, and Yves Scherrer. 2020.
\newblock \href {https://doi.org/10.1017/S1351324920000492} {Natural language
  processing for similar languages, varieties, and dialects: A survey}.
\newblock \emph{Natural Language Engineering}, 26(6):595–612.

\bibitem[{Zhang et~al.(2020)Zhang, Bai, Zhang, Bai, Zhu, and
  Zhao}]{zhang-etal-2020-demographics}
Guanhua Zhang, Bing Bai, Junqi Zhang, Kun Bai, Conghui Zhu, and Tiejun Zhao.
  2020.
\newblock \href {https://doi.org/10.18653/v1/2020.acl-main.380} {Demographics
  should not be the reason of toxicity: Mitigating discrimination in text
  classifications with instance weighting}.
\newblock In \emph{Proceedings of the 58th Annual Meeting of the Association
  for Computational Linguistics}, pages 4134--4145, Online. Association for
  Computational Linguistics.

\bibitem[{Zhou et~al.(2021)Zhou, Sap, Swayamdipta, Choi, and
  Smith}]{zhou-etal-2021-challenges}
Xuhui Zhou, Maarten Sap, Swabha Swayamdipta, Yejin Choi, and Noah Smith. 2021.
\newblock \href {https://doi.org/10.18653/v1/2021.eacl-main.274} {Challenges in
  automated debiasing for toxic language detection}.
\newblock In \emph{Proceedings of the 16th Conference of the European Chapter
  of the Association for Computational Linguistics: Main Volume}, pages
  3143--3155, Online. Association for Computational Linguistics.

\end{thebibliography}
\bibliographystyle{acl_natbib}

\appendix
\section{Dataset Details}\label{dataset}
Our collected dataset is demographically-aligned on AAE in correspondence on the dialectal tweet corpus by \citet{DBLP:journals/corr/BlodgettGO16}. The TwitterAAE corpus is publicly available and can be downloaded from link\footnote{\url{http://slanglab.cs.umass.edu/TwitterAAE/}}. \citet{DBLP:journals/corr/BlodgettGO16} uses a mixed-membership demographic language model which calculates demographic dialect proportions for a text accompanied by a race attribute—African America, Hispanic, Other, and White in that order. The race attribute is annotated by a jointly inferred probabilistic topic model based on the geolocation information of each user and tweet. Given that geolocation information (residence) is highly associated with the race of a user, the model can make accurate predictions. However, there a a low number messages that possess a posterior probabilities of NaN as these are messages that have no in-vocabulary words under the model.


\section{Annotator Annotation Guidelines}\label{sec:human}

You will be given demographically-aligned African American tweets, in which we refer to these tweets as sequences. As a dominant AAE speaker, who identifies as bi-dialectal, your task is to correctly identify the context of each word in a given sequence in hopes to address the issues of lexical, semantic and syntactic ambiguity.

\begin{enumerate}
    \item Are you a dominant AAE speaker?
    
    \item If you responded ``yes'' above, are you bi-dialectal? 
    
    \item If you responded ``yes'', given a sequence, have you ever said, seen or used any of these words given the particular sequence?
    
    \item Given a sequence, what are the SAE equivalents to the identified non-SAE terms?
    
    \item For morphological and phonological (dialectal) purposes, are these particular words spelt how would you say or use them? 
    
    \item If you responded ``no'' above, can you provide a different spelling along with its SAE equivalent? 
\end{enumerate}

\subsection{Annotation Protocol}\label{sec:human_pos}

\begin{enumerate}
    \item  What is the context of each word given the particular sequence? 
    
    \item Given NLTK's Penn Treebank Tagset\footnote{\url{https://www.guru99.com/pos-tagging-chunking-nltk.html}}, what is the most appropriate POS tag for each word in the given sequence?
\end{enumerate}

\subsection{Human evaluation of POS tags Protocol}

\begin{enumerate}
    \item Given the tagged sentence, are there any misclassified tags?
    
    \item  If you responded ``yes'' above, can you provide a different POS tag, and state why it is different?
\end{enumerate}

\section{Variable Rules Examples}\label{sp}

In this section we present a few examples of simple, deterministic phonological and morphological language features or \textit{current} variable rules which highlight several regional varieties of AAE which typically attain misclassified POS tags. Please note that a more exhaustive list of these rules is still being constructed as this work is still ongoing. Below are a few variable cases (MAE $\rightarrow$ AAE), some of which may have been previously shown in Table~\ref{tab:category}: 
\begin{enumerate}

    \item Consonant (`\textbf{t}') deletion (Adverb case) : e.g. ``\textit{just}'' $\rightarrow$ ``\textit{jus}''; ``\textit{must}'' $\rightarrow$ ``\textit{mus}''
    
    
    \item Contractive negative auxiliary verbs replacement: ``\textit{doesn't}'' $\rightarrow$ ``\textit{don't}''

    \item Contractive (\textbf{'re}) loss: e.g. ``\textit{you're}'' $\rightarrow$ ``\textit{you}''; ``\textit{we're}'' $\rightarrow$ ``\textit{we}''

    \item Copula deletion: Deletion of the verb ``\textbf{be}'' and its variants, namely ``\textbf{is}'' and ``\textbf{are}'' e.g. ``\textit{He} \textbf{is} \textit{on his way}'' $\rightarrow$ ``\textit{He on his way}''; ``\textit{You} \textbf{are} \textit{right}'' $\rightarrow$ ``\textit{You right}''
    

    
    \item  Homophonic word replacement (Pronoun case): e.g. ``\textit{you're}'' $\rightarrow$ ``\textit{your}''
    
    
    \item Indefinite pronoun replacement: e.g. ``\textit{anyone}'' $\rightarrow$ ``\textit{anybody}''; 
    
    \item Interdental fricative loss (Coordinating Conjuction case): e.g. ``\textit{this}'' $\rightarrow$ ``\textit{dis}''; `\textit{that}' $\rightarrow$ `\textit{dat}''; ``\textit{the}'' $\rightarrow$ ``\textit{da}''
    
        
    \item Phrase reduction (present/ future tense) $\Rightarrow$ word (Adverb case): e.g. ``\textit{what's up}'' $\rightarrow$ ``\textit{wassup}''; ``\textit{fixing to}'' $\rightarrow$ ``\textit{finna}''
    
    
    \item Present tense possession replacement: e.g. ``\textit{John \textbf{has} two apples}'' $\rightarrow$ ``\textit{John \textbf{got} two apples}''; ``\textit{The neighbors \textbf{have} a bigger pool}'' $\rightarrow$ ``\textit{The neighbors \textbf{got} a bigger pool}''
    
    \item Remote past ``\textit{\textbf{been}}'' + completive (`\textbf{done}'):
    ``\textit{I've \textbf{already done} that}'' $\rightarrow$ ``\textit{I \textbf{been done} that}''
    
    \item Remote past ``\textit{\textbf{been}}'' + completive (`\textbf{did}'): ``\textit{She \textbf{already did} that}'' $\rightarrow$ ``\textit{She \textbf{been did} that}''
    
    \item Remote past ``\textit{\textbf{been}}'' + Present tense possession replacement:
    ``\textit{I \textbf{already have} food}'' $\rightarrow$ ``\textit{I \textbf{been had} food}''; ``\textit{You \textbf{already have} those shoes}'' $\rightarrow$ ``\textit{You \textbf{been got} those shoes}''

    
    \item Term-fragment deletion: e.g. ``\textit{brother}'' $\rightarrow$ ``\textit{bro}''; ``\textit{sister}'' $\rightarrow$ ``\textit{sis}''; ``\textit{your}'' $\rightarrow$ ``\textit{ur}''; ``\textit{suppose}'' $\rightarrow$ ``\textit{pose}'';
    ``\textit{more}'' $\rightarrow$ ``\textit{mo}''
    
    \item Term-fragment replacement: ``\textit{something}'' $\rightarrow$ ``\textit{sumn}''; ``\textit{through}'' $\rightarrow$ ``\textit{thru}''; ``\textit{for}'' $\rightarrow$ ``\textit{fa}''; ``\textit{nothing}'' $\rightarrow$ ``\textit{nun}''
    
\end{enumerate}

\end{document}